\documentclass[sort&compress,natbib,smallcondensed]{article} 

\usepackage{a4wide}
\usepackage[british]{babel}
\usepackage{prettyref}
\usepackage{multirow}
\usepackage{SIunits}
\usepackage{url}

\newrefformat{sec}{section~\ref{#1}}
\newrefformat{fig}{fig.~\ref{#1}}
\newrefformat{tab}{table~\ref{#1}}
\newrefformat{eq}{eq.~\ref{#1}}

\usepackage{amsmath}
\usepackage{pst-plot}
\usepackage{subfig}
\usepackage{epsfig}

\usepackage{mdwlist}
\usepackage{natbib}
\bibpunct{[}{]}{,}{n}{,}{,} 

\usepackage{acronym}
\usepackage{paralist}

\usepackage{svninfo}
\svnInfo $Id: a2.tex 272 2011-12-01 14:32:01Z ag0015 $

\usepackage{fancyhdr}
\acrodef{LTP}{Long-Term Potentiation}
\acrodef{LTD}{Long-Term Depression}
\acrodef{BIMA}{Biolgically Inspired Modelling and Applications}
\acrodef{PI}{Principal Investigator}
\acrodef{RA}{Research Assistant}
\acrodef{BP}{Back-Propagation}
\acrodef{STDP}{Spike-Timing-Dependent Plasticity}
\acrodef{BPTT}{Back-Propagation Through Time}
\acrodef{NN}{Neural Network}
\acrodef{WP}{Work Package}
\acrodef{LSTM}{Long Short-Term Memory}
\acrodef{ICANN}{International Conference on Artifiical Neural Networks}
\acrodef{IJCNN}{International Joint Conference on Neural Networks}
\acrodef{CNS}{Computational Neuroscience Meeting}
\acrodef{DE}{Digital Ecosystems}
\acrodef{FMS}{Formal Methods and Security}
\acrodef{FEPS}{Faculty of Engineering and Physical Sciences}
\acrodef{KTA}{Knowledge Transfer Account}
\acrodef{NCAF}{Natural Computing Applications Forum}
\acrodef{RES}{Research and Enterprise Support}
\acrodef{TTO}{Technology Transfer Office}
\acrodef{LSM}{Liquid State Machine}
\acrodef{ESN}{Echo State Network}
\acrodef{LIF}{Leaky-Integrate-and-Fire}
\newcommand{\rsm}{ReSuMe}
\acrodef{STE}{Spike Train Error}
\acrodef{LE}{Logic Error}
\acrodef{ISI}{Inter-Spike Interval}

\newcommand{\true}{\textsc{true}}
\newcommand{\false}{\textsc{false}}
\newcommand{\XOR}{\emph{XOR}}
\newcommand{\AND}{\emph{AND}}
\newcommand{\TRUE}{\emph{TRUE}}
\newcommand{\PROJ}{J0}
\newcommand{\J}{\PROJ}
\newcommand{\OR}{\textsc{OR}}

\title{Supervised Learning of Logical Operations in Layered Spiking
  Neural Networks with Spike Train Encoding}

\author{Andr\'e Gr\"uning and Ioana Sporea \\
  Department of Computing, University of Surrey, \\
  Guildford, GU2\,7XH, United Kingdom, 
  \url{a.gruning@surrey.ac.uk}}

\begin{document}

\maketitle

\begin{abstract}
Few algorithms for supervised training of spiking neural networks exist 
that can deal with patterns of multiple spikes, and their
computational properties are largely unexplored. We  
demonstrate in a set of simulations that the \rsm{} 
learning algorithm can be successfully applied to layered neural
networks. Input and output patterns are encoded as spike trains of
multiple precisely timed spikes, and the network learns to transform
the input trains into target output trains. This is done by combining
the \rsm{} learning algorithm with multiplicative scaling of the
connections of  downstream neurons.   
\par
We show in particular that layered networks with one hidden layer can
learn the basic logical operations, including Exclusive-Or, while
networks without hidden layer cannot, mirroring an analogous result
for layered networks of rate neurons.  
\par
While supervised learning in spiking neural networks is not yet fit
for technical purposes, exploring computational properties of spiking
neural networks advances our understanding of how computations 
can be done with spike trains.

\end{abstract}

\acresetall

\section{Introduction}

Artificial neural networks are developed both as models of neural
processing in nervous systems and as learning devices in artificial
intelligence.  Neural networks of \emph{rate neurons} have found ample 
applications in industry because efficient general purpose
learning algorithms exist.  In our understanding, a general purpose
algorithm can learn arbitrary mappings of input-output pattern pairs,
-- subject to general constraints of whether the input-output mapping
is representable in the network or crosstalk between similar input patterns \citep{legenstein:05a, bengio:94}. Examples of such general-purpose algorithms for rate neurons are the 
family of backpropagation algorithms \citep{rumelhart:86}.  However networks of rate neurons are not biologically
plausible as they do not show  spiking behaviour.  
\par
On the other hand, spiking neural networks are biologically
more plausible and serve mainly as models of nervous processing, but
general purpose learning algorithms -- in the way  
backpropagation is applied to rate neurons -- have not yet been found
\citep{kasinski:06}.  A general-purpose learning algorithm for spiking networks
should be able to map arbitrary spatio-temporal input spike patterns to
arbitrary output spike patterns.  So far learning in these spiking networks is largely
correlation-based, that is variants of Hebbian learning such as \ac{STDP}
are typically used to change synaptic weights \citep{kistler:00}. 
\par
In this paper we will present a series of simulations involving
layered feedforward networks of spiking neurons and demonstrate that
these are able to learn simple computations in a supervised way. We will use an encoding
of input and output patterns that makes use of spike trains with
strict spike times. In comparable settings, so far only classification
tasks or simple mapping tasks have been considered
\citep{izhikevich:07, legenstein:08, ponulak:10, kasinski:05}, either with only
a single neuron or in much larger \acp{LSM} \citep{maass:01}, but no
computational tasks. Or computational tasks like the Exclusive-Or problem
have been considered in layered networks, but only with single-spike
latency-encoded outputs \cite{bohte:02, booij:05, tino:xx, sporea:11}.
In contrast, it is our aim to demonstrate that layered networks can
learn to perform simple, but non-trivial computations in a supervised framework and make use of
multiple timed spikes for input and output patterns.
\par
In particular, as basic building blocks of computation, we demonstrate
that these networks can learn logical operations when logical values
\false{} and \true{} are encoded as spike trains both for
inputs and outputs. While it can often be shown that 
(hand-coded) spiking neural networks can be Turing-equivalent
\citep{maass:01, siegelmann:92}, it is instructive to demonstrate that
basic building blocks of such computations can indeed be
learnt. Already for rate neurons, theoretical Turing equivalence and
practical learnability of a problem may not coincide \citep{boden:00b,
  gruening:05}.  
\par
A key problem in neuroscience is to understand the neural
code. Usually the approach is ``bottom-up'', that is spike 
trains are recorded and later analysed and checked for correlations
in rate and timing with experimental conditions, for example sensory
stimuli \citep{nemenman:08}. However many areas of a nervous system
might be so far remote from direct sensory stimuli, that it is
difficult to detect such correlations and to understand what precise
computational function a natural network implements and how. 
\par
Therefore, besides as an initial step towards general-purpose learning
for spiking neural networks, the present article may also be seen as a
top-down complement of these neuroscientific approaches. Under
biologically inspired constraints on information processing, we
explore whether simple types of computation can be performed with
spike-train based encoding.  While bio-inspired, our approach however
takes into account neuroscientific detail on a coarse level only. 
\par 
The article is structured as follows: In the next
\prettyref{sec:background} we discuss basic properties of two learning  
algorithms for spiking neurons and motivate our choice of \rsm{}. We present our learning task in
\prettyref{sec:task}. In \prettyref{sec:methods} we describe the
details of the simulation setup.  Section~\ref{sec:sim} presents and discusses a series
of simulations on logical operations. In \prettyref{sec:conclusion} we
conclude by embedding the results into their wider context.

\section{Background} \label{sec:background}

Recently there have been interesting developments regarding supervised
learning algorithms for spiking neural networks. Notably, there are
the SpikeProp learning algorithm \citep{bohte:02,booij:05} and
\rsm{} \citep{ponulak:10}. 
\par
While SpikeProp has been applied to layered feedforward networks, each
neuron is restricted to only one spike during a certain
period. Similar restrictions also apply to extensions of this
algorithm \citep{tino:xx, booij:05}. SpikeProp essentially is a
gradient-descent algorithm 
similar to backpropagation for rate neurons. While rate neuron
backpropagation uses the minimisation of Euclidean distance between actual and target
output \emph{activation} to derive weight changes to minimise error,
in SpikeProp the Euclidean distance of actual and target spike \emph{times}
plays the same role. As in standard backpropagation, SpikeProp
weight changes for synaptic connections between neurons are given by
the (anti-)gradient of the overall network error with respect to
the weight. Such gradient descent algorithms overcome the credit
assignment problem by utilising the chain rule
of differentiation to derive error signals for downstream neurons.
\par
Applications of SpikeProp and its extensions have mainly been to
classification tasks in layered networks where the early or late timing
of a single output spike indicates the class \citep{booij:05, tino:xx,
schrauwen:xx}. SpikeProp's
application to the non linearly separable Exclusive-OR problem also
follows this pattern \citep{bohte:02}. Generally for 
SpikeProp-based algorithms, it is crucial that hidden layer neurons
are initialised such that they spike at least once for all 
patterns or no error signals for that neuron and its weight arise.
In this sense, it is difficult to come up with a good weight initialisation
independent of the task and the pattern encoding used
\citep{schrauwen:xx}.    
\par
Finally, \citet{booij:05} suggest an extension of SpikeProp where
multiple spikes are allowed in the hidden and input layers. They claim their extension is in
principle also applicable for multiple output spikes, but -- to our
knowledge -- it has never been successfully applied to any such
task experimentally. Also our own preliminary simulations with
SpikeProp failed for multiple output spikes. Therefore simulations in the paper will be based on
\rsm{}, see \prettyref{sec:learning}.
\par
\rsm{} is motivated by an analogue of the $\delta$-rule and is based
on \ac{STDP} \citep{ponulak:10}. From a combination of an \ac{STDP}
process between target 
and input spike trains of a neuron and an anti-\ac{STDP}
process between the actual output train and the input, it derives a
differential \ac{STDP} rule that is used to generate weight
changes for the synaptic connections into a neuron. 
\par
This algorithm is capable of training a single neuron to reproduce an
arbitrary prescribed target spike train via supervised learning;
however it needs a large number of incoming   
spikes to do so successfully \citep{ponulak:10}. Unlike SpikeProp, \rsm{} is able to deal with spike patterns that involve many
spikes. However it can only be applied to neurons that have a
direct target spike train assigned, and the credit assignment for
downstream neurons is circumvented by either training single neurons
or a layer of output neurons on top of a large immutable \ac{LSM}
\citep{maass:01}. Experimental tasks include, for 
example, single neurons (or sets) producing a prescribed spike train
from their incoming spikes, or classifying spike patterns when
trained neurons are used as readouts for a \ac{LSM}
\citep{ponulak:10}. 
\par
However, much smaller networks, similar to the layered feedforward
networks used for SpikeProp can also perform computations and
transform spike trains as will be demonstrated 
in this article. Hence \rsm{} has the potential to be a
general-purpose learning algorithm for patterns that are based on
(arbitrary) spike trains.

\section{Task Overview} \label{sec:task}

This section is an overview of the learning task, the encoding and the
network structure used. For details, see the \prettyref{sec:methods}. We concentrate on simple logical
operations . These are simpler than real world data, but it is also
much clearer  what type of computation has to be learnt in the
network. We are primarily interested in these simulations  as a proof
of concept that  computation with spike trains is possible, but logical
operations are also at the heart of every symbolic computation, and it
is instructive to analyse whether these basic building blocks can be
learnt.   
\par
\begin{figure} 
  \centering
  \subfloat[Network
  structure.]{\epsfig{width=0.5\hsize,file=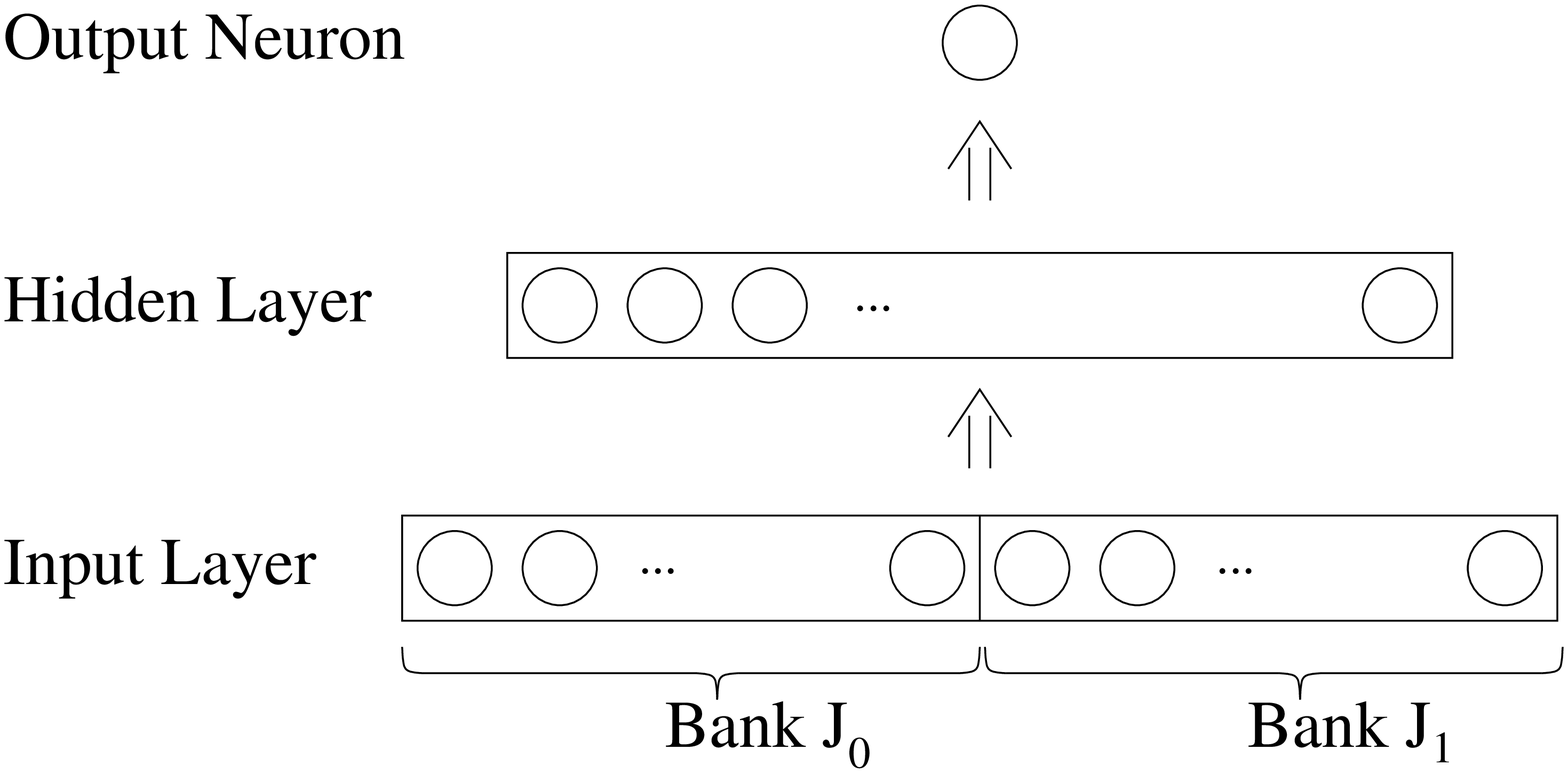}} \\
  \subfloat[Logical Operation.]{\epsfig{width=0.9\hsize,file=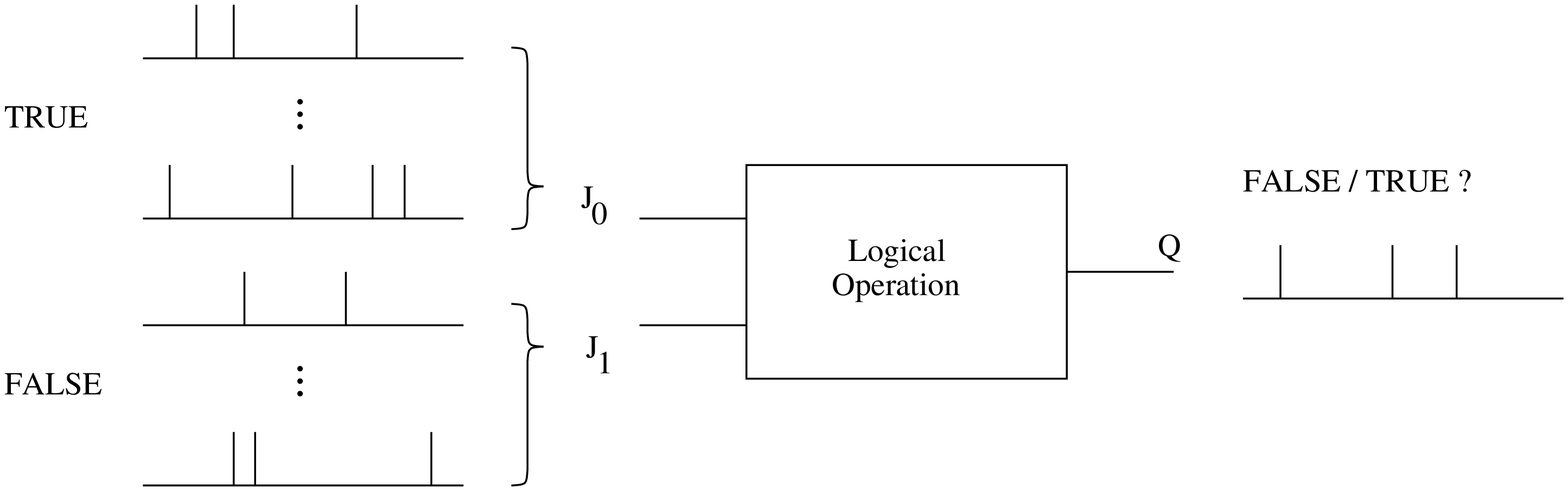}}
  \caption{Network structure and encoding of logical values. (a) A
    three-layer network with input, hidden and output layer. The
    input layer consists of two banks $J_0$ and $J_1$. All neurons in
    a bank collectively serve as one logical input. The firing pattern
    of neuron $Q$ 
    represents the output value. $\Uparrow$ stands for the
    feedforward connections between layers. 
    (b) Logical operations with spike trains. A logical value \false{} or \true{} is
    encoded in a set of spike trains, and these are applied to the
    input banks $J_0$ and $J_1$. The network learns to perform a
    logical operation (such as \XOR{}) and produces an output spike 
    train which then needs to be interpreted as a logical value.}
  \label{fig:network}
\end{figure}

Let $J_0$ and $J_1$ denote the inputs to a logical operation and $Q$ its
output. Truth values \false{} and \true{} both for input and output
will be encoded as spike trains for a layered feedforward network
\citep{rojas:96} of spiking neurons (see \prettyref{fig:network}).  For the single output neuron $Q$ (in slight abuse of
notation), there are two target spikes trains  $S_\true$ and
$S_{\false}$, standing for the two logical output values.  For the
inputs, the network has two equally sized banks $J_0$ and $J_1$ of
input neurons. Each bank plays the role of one logical input to the
network. For each bank $J_i$ the list of given  spike trains for the
bank's individual neurons~$(S)_{J_i,  \true}$ or $(S)_{J_i,\false}$
denote collectively the logical value input to this bank. For details
of spike train choice, see  \prettyref{sec:trains}.
\par
We have chosen to train the four operations \TRUE, \J, \AND{} and
\XOR. As spike trains are assigned randomly to their interpretations
of \false{} and \true{} and to their bank, but have otherwise
identical properties, these four cases cover all 16 possible logical
operations of two binary variables \citep{enderton:01}. For example \OR{} can be derived
from \AND{} if logical values of all input and target spike trains are
inverted.   \TRUE{} and \J{} might seem trivial from a logic point of
view, but they are probably not for a spiking neural network: 
\begin{description*}
\item{\TRUE} is the logical operation that always has \textsc{true} as output,
  irrespective of its inputs. It tests whether the network can produce the
  same (or at least a similar) output train for dissimilar sets of input trains.
\item{\J} is the logical operation that always has the same value as
  its $J_0$ input. It tests whether the network can ignore the input
  from bank $J_1$ which effectively is just noise in this task. 
\item{\AND}. The logical conjunction is linearly separable, can therefore be
  learned in a single layer preceptron, and is viewed as a simple
  computation to learn.
\item{\XOR}. The Exclusive-Or operation is not linearly separable and
  is therefore considered more difficult to learn than AND. It cannot be
  learnt with a simple preceptron and has frequently been used to
  demonstrate the power of a learning algorithm \citep{minsky:88}
\end{description*}

\section{Methods} \label{sec:methods}

In this section we describe and motivate our experimental setup in
more detail. 
\par
\rsm{} is a supervised learning algorithm for single spiking neurons
usually driven by a large number of input spike trains
\citep{ponulak:10}. \rsm{} is not fixed to a particular neuron type,
but -- as \ac{STDP} -- implicitly assumes, at least on longer time
scales, that recent inputs have more influence on the current
activation of a neuron than past inputs.  We present the general
network structure,  fix a convenient notation and introduce \rsm{} in a form
suitable for easy implementation with discrete time steps as opposed
to the integral formulation  in continuous time  in
\citep{ponulak:10}. We also address the
problem of weight initialisation for downstream neurons.

\subsection{Networks} \label{sec:network}

We consider layered feedforward networks \citep{rojas:96}. Although they are only
feedforward and not recurrent, they have a temporal dimension since
they use spiking neurons and spike train dynamics play out in
time. Our simulations used networks with two 
(input, output) and three layers (input, hidden and output). The
two-layered networks are similar to the ones used for \rsm{}
\citep{ponulak:10} and the three layered ones are similar to the ones
used with SpikeProp \citep{bohte:02}.
\par
Neurons within layers do not connect to each other, but fully
connect to the subsequent layer: there are multiple connections $w_{XY,s}$
for all delays $s =1\milli\second \cdots 10\milli\second$ from any neuron $X$ in the present
layer to any neuron $Y$ in the subsequent layer. The output layer
consists of just a single neuron, and the input layer 
consists of ``dummy'' neurons (without any dynamics) which simply serve to feed input
spike trains into the next layer.
\par
The connections between the input and output layer for the two-layered
networks are subject to \rsm{} learning (\prettyref{eq:dwd} and \prettyref{eq:dwa}). 
In the three-layer network only the connections from the hidden layer
to the output layer are subject to \rsm{} learning, while
connections from the input layer to the hidden layer are subject to
rate adjustment according to \prettyref{eq:adj}, further down.  
\par
\rsm{} needs a large number (hundreds) of incoming connections to
function \citep{ponulak:10}, we (like in SpikeProp) instead use fewer
inputs, but multiply incoming connections by having 10 weights with
different delays between any two neurons, so effectively we also
achieve a high number of incoming spikes to any neuron.

\subsection{Notation}

Let $S_X$ denote the output spike train from neuron $X$. We understand
the spike train as the ordered set of spike times $t_{i}$ of $X$, ie $S_X = (t_{i})$. A neuron $X$ undergoing 
supervised learning will have \emph{two} output spike trains
associated with it, namely the train of \emph{actual} spikes
$S_X^{(a)} = (t_{i}^{(a)})$, that is the list of times $t_i^{(a)}$ when it
did actually spike, and the train $S_X^{(d)} = 
(t_{i}^{(d)})$ of desired spike times $t_{i}^{(d)}$ when we
want it to spike. Let $w_{XY}$ denote the
weight of the synaptic connection from presynaptic neuron $X$ to
postsynaptic neuron $Y$. We distinguish multiple synaptic connections between the same neurons~$X$ and $Y$ with different delays $s$ by an additional index, that is $w_{XY,s}$. If clear from the  
context which particular quantity we refer to or if the
argumentation is generic, we will leave out indices on weights
$w_{XY}$, times $t_{i}$ and on other quantities introduced later.

\subsection{Weight Changes in \rsm{}} \label{sec:learning}

\rsm{} considers a single neuron $Y$ that is driven by a number of
incoming spike trains, either 
as direct input spike trains or trains from other neurons. It
introduces a differential \ac{STDP} process involving the desired and
actual output spike trains $S^{(d)}_Y$ and $S^{(a)}_Y$ and all input
spike trains. We refer the reader to details of 
its derivation in \citep{ponulak:10} and present a formulation
of \rsm{} that is broken down to the effects of individual
input-output spike pairs. 
\par
More precisely, for each connection $w_{XY}$ there is an \ac{STDP} process
between the corresponding input train  $S_X$ and the desired output
train $S_Y^{(d)}$. This process is complemented by an anti-\ac{STDP}
process between the same incoming train $S_X$ and the actual output
train $S_Y^{(a)}$. These processes can be formulated quite generally with a
number of parameters, however, we restrict ourselves here to a
formulation with a reduced set of parameters where contributions of
the two \ac{STDP} processes are of equal magnitude. 
\par
The total weight change $\Delta w_{XY}$ resulting from the
\ac{STDP} processes between the trains is the sum of all contributions
of individual input-output spike pairs $(t_{X}, t^{(d)}_{Y}) \in S_X
\times S_Y^{(d)}$ and $(t_{X}, 
t^{(a)}_{Y})\in S_X \times S_Y^{(a)}$  as follows: 
\begin{equation}
  \Delta w_{XY}^{(d)}(t_{X},t_{Y}^{(d)}) = a_d + 
  \begin{cases}
    + A_{di} e^{-\frac{t_{Y}^{(d)} - t_{X}}{\tau}}, & t_{Y}^{(d)} - t_{X} \geq 0 \\
    - A_{id} e^{\frac{t_{Y}^{(d)} - t_{X}}{\tau}}, & t_{Y}^{(d)} - t_{X} < 0
  \end{cases} \label{eq:dwd}
\end{equation}
with constants $a_d, A_{di}, A_{id}, \tau \geq 0$. Similarly, the anti-\ac{STDP} process between an input train
and the \emph{actual} output train effects weight changes as
\begin{equation}
  \Delta w_{XY}^{(a)}(t_{X},t_{Y}^{(a)}) = -a_d +
  \begin{cases}
    - A_{di} e^{-\frac{t_{Y}^{(a)} - t_{X}}{\tau}}, & t_{Y}^{(a)} - t_{X} \geq 0 \\
    + A_{id} e^{\frac{t_{Y}^{(a)} - t_{X}}{\tau}}, & t_{Y}^{(a)} - t_{X} < 0
  \end{cases}.  \label{eq:dwa}
\end{equation} 
With constants chosen the same as in \prettyref{eq:dwd}, $\Delta w^{(a)}(t_X, t_Y) =
- \Delta w^{(d)}(t_X, t_Y)$ so that the two processes are balanced.
If the desired and actual spike times coincide, there is no further weight change resulting from such a
pair. The total weight change of $w_{XY}$ from spike trains $S_X,
S_Y^{(a)}$ and $S_Y^{(d)}$ is the sum of all above contributions from
all pairings of (desired and actual) output and input spikes:
\begin{equation} 
  \Delta w_{XY} = \sum_{t_{X} \in S_X} \left(
  \sum_{t^{(d)} \in S^{(d)}_Y}   \Delta w_{XY}^{(d)}(t_{X},t^{(d)}) +
  \sum_{t^{(a)} \in S^{(a)}_Y}   \Delta w_{XY}^{(a)}(t_{X},t^{(a)}) \right) \label{eq:ponu29}
\end{equation}
If connections $w_{XY,s}$ have delays $s$ then in the above formulas
$t_X + s$ replaces $t_X$. 
\par
Learning parameters used were $a_d = 0$, $A_{di} = 0.0005, A_{id} =
A_{di}$ and $\tau = 4\milli\second$ in all cases. Preliminary
simulations had shown that for these values we could expect a
reasonable convergence of networks with three layers and that higher
rates led to no stable convergence.  
\par
Note that in practice often $A_{id} = 0$ so that \prettyref{eq:dwd} and
\prettyref{eq:dwa} only yield a non-zero contribution  
for those presynaptic spikes $t_X$  that arrive before the current
desired or actual spike $t_Y$ considered
\citep{ponulak:08}. Preliminary simulations in our setting
showed that $A_{id} = 0$ did not work well, presumably because 
$A_{id} > 0$ makes most difference for a connection $w_{XY}$ when for the incoming
spike $t_X$ either $t_Y^{(a)} < t_X < t_Y^{(d)}$ or $t_Y^{(a)} > t_X >
t_Y^{(d)}$, because in these cases \prettyref{eq:dwd}
and~\ref{eq:dwa} have the same sign.

\subsection{Adjusting Spike Rates for Downstream Neurons}

We suggest a general natural method to overcome the 
problem of silent downstream neurons, that can 
hamper learning of upstream neurons. 
\par
In SpikeProp many problems arise because neurons in the hidden layer
do not fire, and it is not straight forward how to overcome this,
other than by careful selection of initial weights, so that all
neurons fire at least one spike for all input patterns to the
network. Thus weight initialisation depends on the task
\citep{schrauwen:xx}. Our layered network has in principle
the same problem. Although connections to the  neurons in the hidden
layer are not actively trained in the \rsm{} sense, firing of hidden
neurons needs to be tuned to produce a sufficient number of incoming
spikes for the output layer.  
\par
Natural neurons can multiplicatively scale incoming
synapses collectively to keep their output firing rate within an
acceptable range \citep{shepard:09}. This natural scaling is adopted
into our network: If we set a target spike rate range
$[r_{min},r_{max}]$ for a neuron, weights are scaled when neuron  $Y$'s
average rate is outside this range:
\begin{equation}
  \label{eq:adj}
  w_{XY} \to 
  \begin{cases}
    (1 + f) w_{XY}, & w_{XY} > 0 \\
    \frac{1}{1 + f} w_{XY} & w_{XY} < 0
  \end{cases} 
\end{equation}
with $f > 0$ for $r_Y < r_{min}$ and $f < 0$ for $r_Y > r_{max}$.
If the hidden neurons act as preprocessors of the input for the output
neurons, it makes sense to hold their rates roughly between those of
the input spike trains and the desired output spike train (see
below). We set $r_{min} = 0.3/\milli\second$ and $r_{max} = 0.1/\milli\second$ with $f = \pm
0.05$.

\subsection{Neurons and Synapses}
All neurons in the network, except the input neurons, are standard
\ac{LIF} neurons \citep{gerstner:02}:
\begin{gather}
  \frac{dV}{dt} = - \frac{1}{\tau} (V - V_r) + \frac{1}{C} I 
\end{gather}
where $V$ is the current membrane potential, $V_r = -60\milli\volt$ the resting
potential. $C$ is the membrane capacity, and with $R$ the membrane
resistance, $\tau := RC$ is its time constant. Finally $I$ is the input
current. If $V$ 
exceeds $V_\theta = -55\milli\volt$, the neuron fires a spike and the membrane
potential is reset to $V_o = -65\milli\volt$. For simplicity, we do not enforce
an absolute refactory period. Neurons are pulse-coupled through
synapses with a numeric weight $w_{XY,s}$ and a delay $s$, that is if
neuron $X$ reaches the firing threshold $V_\theta$ at $t_X,$ $Y$ gets
a contribution $w_{XY,s}$ to its input current
$I$ at $t_X + s$. 
\par
We simulate the neuron with a time resolution of $\Delta t = 1\milli\second$, and choose $R = 10
\mega\ohm$, $C = 1\nano\farad$, hence $\tau = 10\milli\second$.  If we measure $V$ in $[\milli\volt]$,
$w$ in $[\nano\ampere]$ and times in $[\milli\second]$, then an incoming spike with $w =
1\nano\ampere$ with duration $1\milli\second$ (according to the time
resolution) increases the membrane voltage instantaneously by $\Delta
V = w \Delta t / C = 1 \nano\ampere\usk\milli\second\per\nano\farad =
1\milli\volt$. Hence with this choice of dimensions, the numeric value
of a weight corresponds to the numeric value of the instantaneous
increase of the membrane voltage. We will therefore leave out
dimensions on weights, potentials and times in the following.
\par
All weights $w_{XY,s}$ in the network with delays from $s = 1 \cdots
10$ are initialised uniformly from range $-0.02$ to $0.08$,
deliberately chosen small so that no output spikes are produced until
the \rsm{} learning or scaling \prettyref{eq:adj} have increased the
weights. The distribution is skewed towards positive values to coarsely reflect
distribution of excitatory and inhibitory neurons in the brain, if not
in the type of neuron, at least in the type of connection
\citep{okun:09}. Weight are subject to changes according to 
\prettyref{eq:dwd}, \prettyref{eq:dwa} or \prettyref{eq:adj}, however are clipped to values
within range $[-2,2]$ so that several spikes need to contribute to a
neuron's firing. Weights can change seamlessly from excitatory
(positive values) to inhibitory (negative values) and vice-versa.

\subsection{Spike train} \label{sec:trains}

We create spike trains for inputs and outputs that stand for logical
values \false{} and \true{}. For rate neuron networks it is known
that they frequently fail to discriminate between input patterns that
are too similar. Preliminary simulations in our spike train setting showed
that this is also the case here. In addition, actual output spike
trains tend to have additional or missing spikes compared to desired
spike trains. Therefore to ensure a good degree of dissimilarity
between spike trains for the different logical values we proceed as
follows:  
\par
For each input or output neuron, first a single spike train
$S_0$ is created with constant spike probability per time slot $r =
0.2/\milli\second$ for input trains and $0.06/\milli\second$ for output trains, both with a
minimum \ac{ISI} of 10\milli\second{} (mimicking a refactory period). From train $S_0$ two new trains
$S_{\true}$ and $S_{\false}$ are created by randomly distributing
all spikes from $S_0$ over $S_{\true}$ and $S_{\false}$. This ensures
that for a spike $t \in S_{\true}$ there are no spikes in $S_{\false}$
in the interval $[t - 10, t + 10]$ and vice-versa
\par
Spike train pairs $(S_\true, S_\false)$ of duration 100\milli\second{} for logical
zeros and ones are so created 
independently for all input neurons without any further
constraints. Truth value patterns for an input bank are just the set
of the respective trains for the bank's individual input neurons. For the
output neuron, spike train pairs of 100\milli\second{} duration are created in the same
way, but only those selected that have no spikes within the first 20\milli\second{} and so
that each train $S_\true$ and $S_\false$ has 3
spikes.

\subsection{Epochs and Weight Updates}

One epoch consists of ten input-target pattern pair presentations. 
For each such presentation, we choose randomly logical values for the
two input banks $J_0$ and $J_1$, and apply the corresponding sets of spike
trains to the input neurons. The network runs for the simulated 100\milli\second{} duration
of the input trains plus 20\milli\second{} (two times the maximal synaptic
delay). The output spike train is recorded and, after each presentation, weight changes for all connections between
hidden and output layer are calculated (but not applied) with
\prettyref{eq:dwd} and \prettyref{eq:dwa}. The network is then reset (all neurons 
set to $V_r = 60\milli\volt$), and the next input-target pair selected. 
\par
At the end of an epoch, that is, after each 10 presentations of input-target
pairs, the accumulated weight changes are applied to the weights
between  the hidden and output layer. Also, the average rate of the
hidden neurons over this epoch is checked and weights between the input 
and the hidden layer scaled with \prettyref{eq:adj} if necessary. Finally, in each epoch,
we test the network on all pattern pairs (four for a logical
operation), record the results and calculate two error measures.

\subsection{Error Measures} \label{sec:errors}

Unlike gradient-descent algorithms that start from an explicit error
measure between actual and desired output, for \rsm{} there is no
such natural choice since it starts from a pair of \ac{STDP} processes.
Although \rsm{} is motivated with the $\delta$-rule, this does not
provide an immediate error measure since pairing of actual and desired
output spikes is not obvious or even possible.  Errors in the
simulations are therefore measured as follows: 
\begin{enumerate}
\item \emph{\ac{STE}:} Our primary error measure for the difference between actual and
  desired spike trains is from \citet{rossum:01}. It accounts for
  additional and missing spikes as well as a close match of spike
  times.  Given a spike train~$S$ as an ordered set of spike times, we can
  easily view it as function in time:
  \begin{equation}
    \hat S(t) = \sum_{t' \in S} \delta(t - t')
  \end{equation}
  $\hat S$ is convolved with $f(t) = e^{-t/\tau_c}H(t)$ ($H$ is the
  Heaviside function) where the discrete convolution
  runs over the length of 120\milli\second:
  \begin{equation}
    (f \star \hat S)(t) = \sum_{\forall s: 0
      \leq t-s < 120} f(s)\hat S(t-s), 0 \leq t < 120,
  \end{equation}
  $\tau_c = 10\milli\second$ is in the order
  of the \ac{ISI} of input and target trains. The distance $R$ between
  two spike trains $S,T$ is the squared distance between their
  convolutions  
  \begin{equation}
    R(S,T) := \sum_{0 \leq t < 120} \left( (f \star \hat S)(t) - (f
      \star \hat T)(t) \right)^2
  \end{equation}
  Finally, the \ac{STE} is the sum of the distances $R$ between the actual output
  and the target train for all four test cases.
\item The \emph{\ac{LE}} is the count of wrong outputs:
  We count an output train $S^{(a)}$ as correct if it is closer to the spike train $S_q$ of the target
  logical value~$q$ than to $S_{\neg q}$, that is if
  $R(S^{(a)}, S^{(d)}_q) < R(S^{(a)}, S^{(d)}_{\neg q})$. \ac{LE} is the
  number of output trains in the four test cases that are not correct
  with this criterion, so it ranges from $0$ to $4$. 
\end{enumerate}
\ac{STE} is used to generally measure how closely an actual output spike train
matches its target spike train, while \ac{LE} is our criterion to
decide whether we accept an actual output train as the correct
response of the network, namely when it is closer to the target train
than to the non-target train.

\section{Simulations}  \label{sec:sim}

For each of the four logical operations \XOR, \AND, \J{} and \TRUE, we trained three-layer
networks and two-layer networks with the following configuration:
\begin{enumerate*} 
\item three-layer networks with 2x6 inputs and 20 hidden neurons. \label{nw:12_20}
\item two-layer networks with 2x10 input neurons. \label{nw:20}
\end{enumerate*}
For each configuration and logical operation, 100 networks were run
for 2000 epochs, and each run had a different random weight
initialisation. Each run had also its individual random set of spike
trains for input banks and outputs as described in \prettyref{sec:trains}. 
\par
Our main interest is certainly in networks with three layers that
are trained on \XOR. The other logical operations and network
configurations serve as control cases.

\begin{figure} 
  \centering
  \subfloat[\ac{STE} for \XOR]{\epsfig{width=0.45\hsize,file=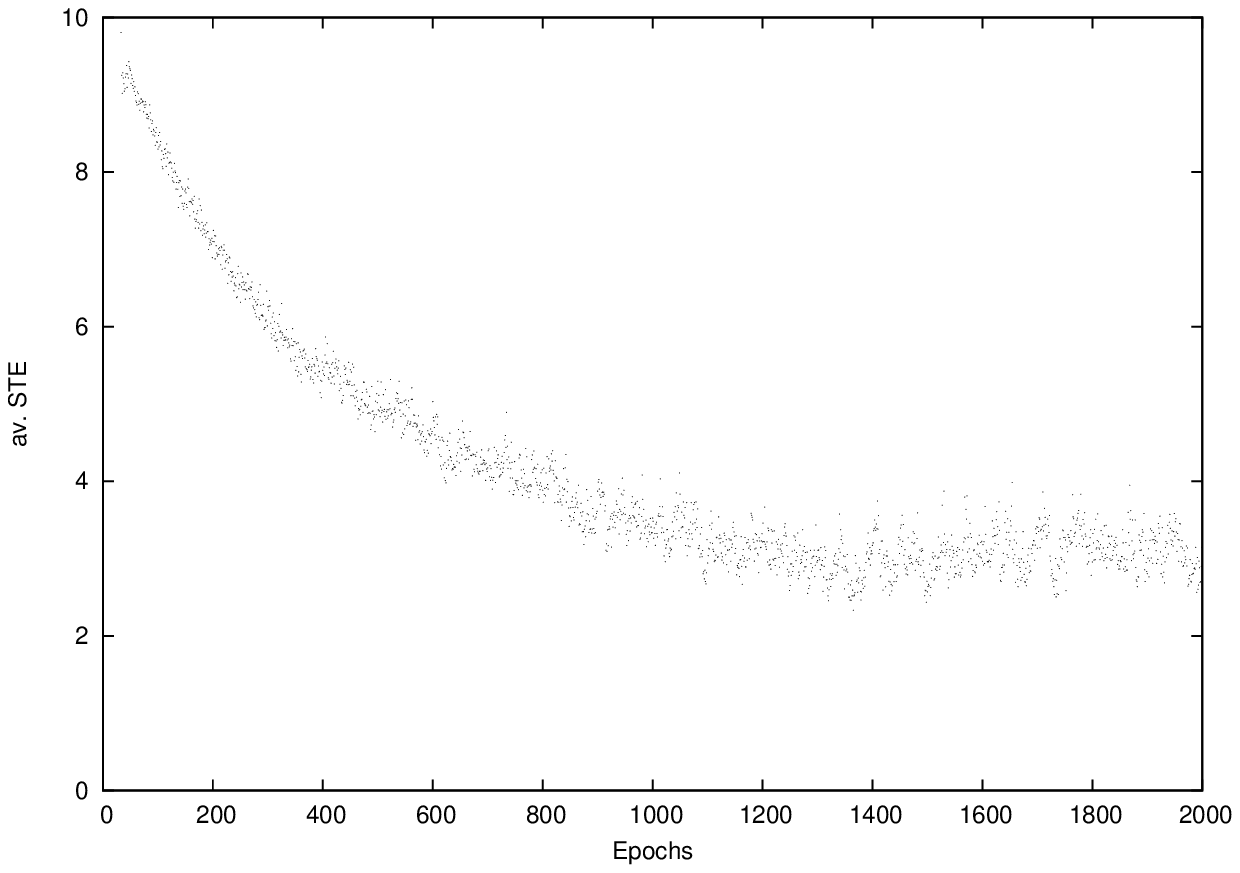}}
  \subfloat[\ac{STE} for \AND]{\epsfig{width=0.45\hsize,file=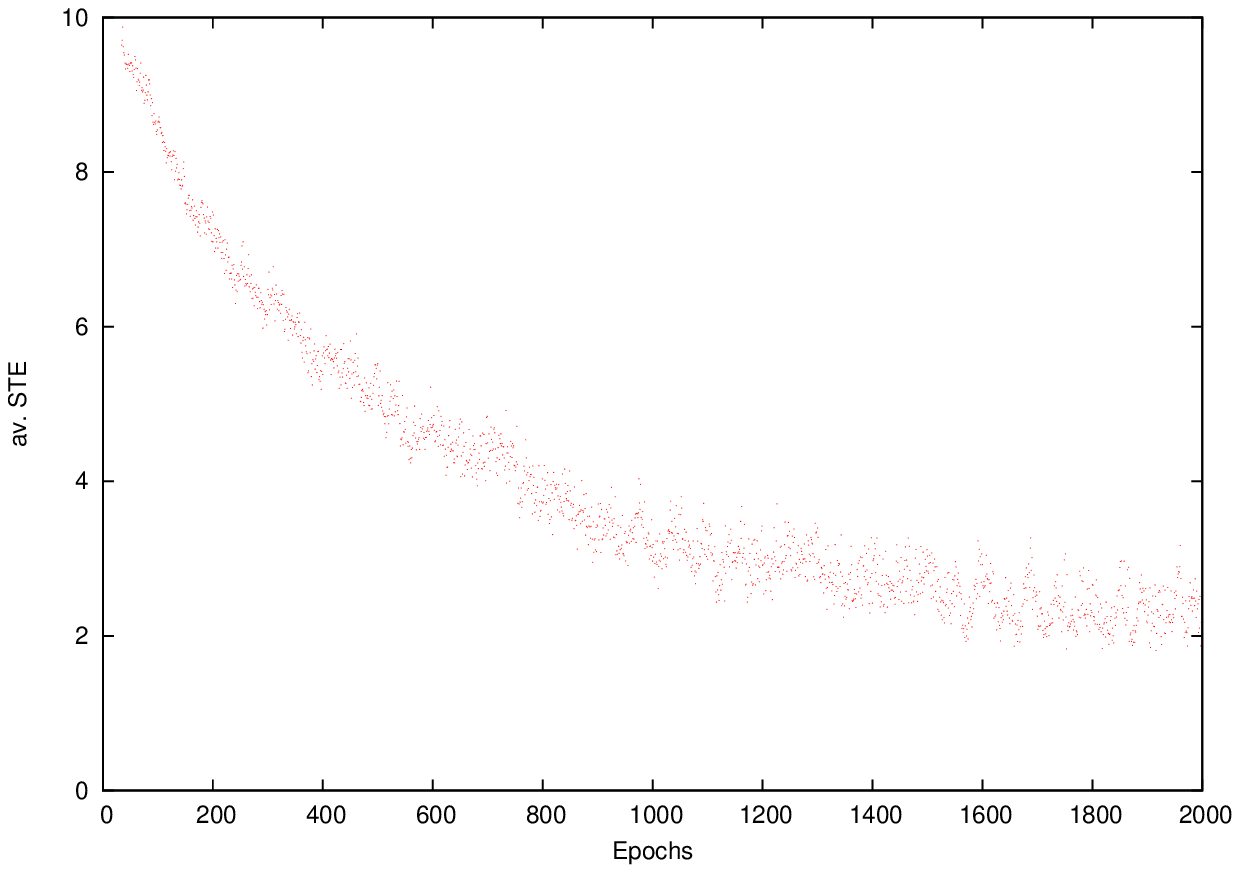}}
  \\

  \subfloat[\ac{LE} for \XOR]{\epsfig{width=0.45\hsize,file=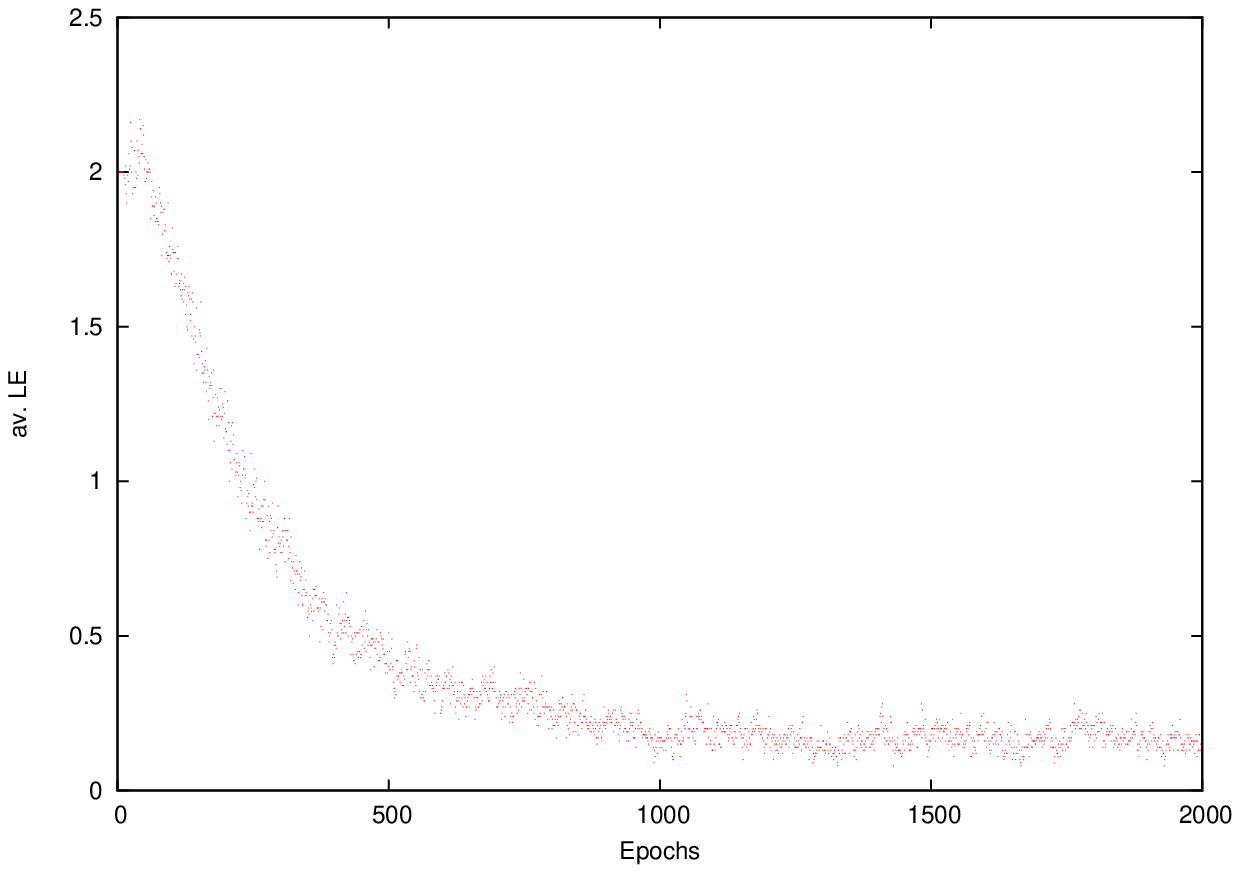}} 
  \subfloat[\ac{LE} for \AND]{\epsfig{width=0.45\hsize,file=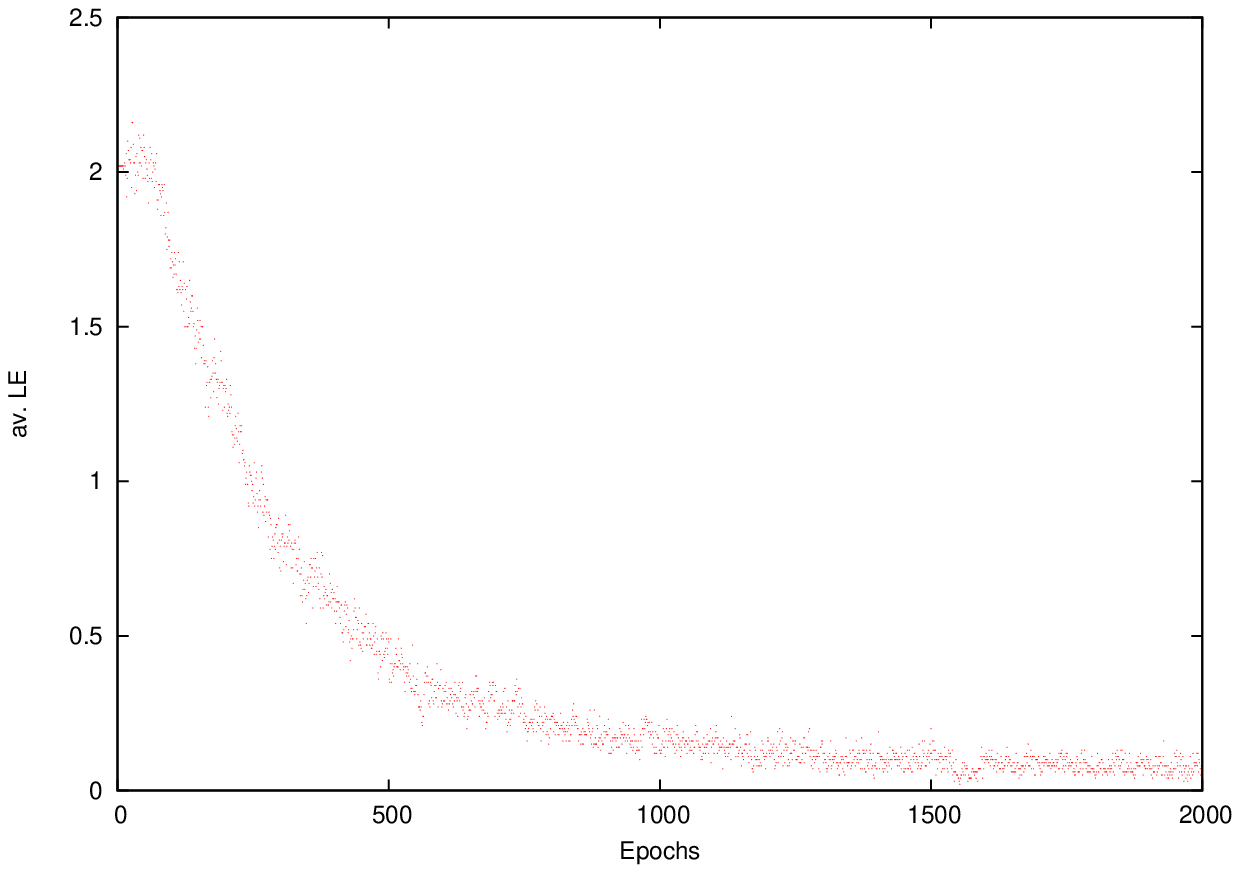}}
  \caption{Average \ac{STE} and \ac{LE} for logical operations \XOR{}
    and \AND{} for the three-layer networks with 20 hidden neurons and
    12 inputs. The \ac{STE} graphs are clipped at 10 as this error is only exceeded for the first
    20 epochs as discussed in the main text.}  \label{fig:av_12_20_ste} \label{fig:av_12_20_le} 

\end{figure}

\begin{figure}
  \centering
  \subfloat[\ac{STE} for \XOR]{\epsfig{width=0.45\hsize,file=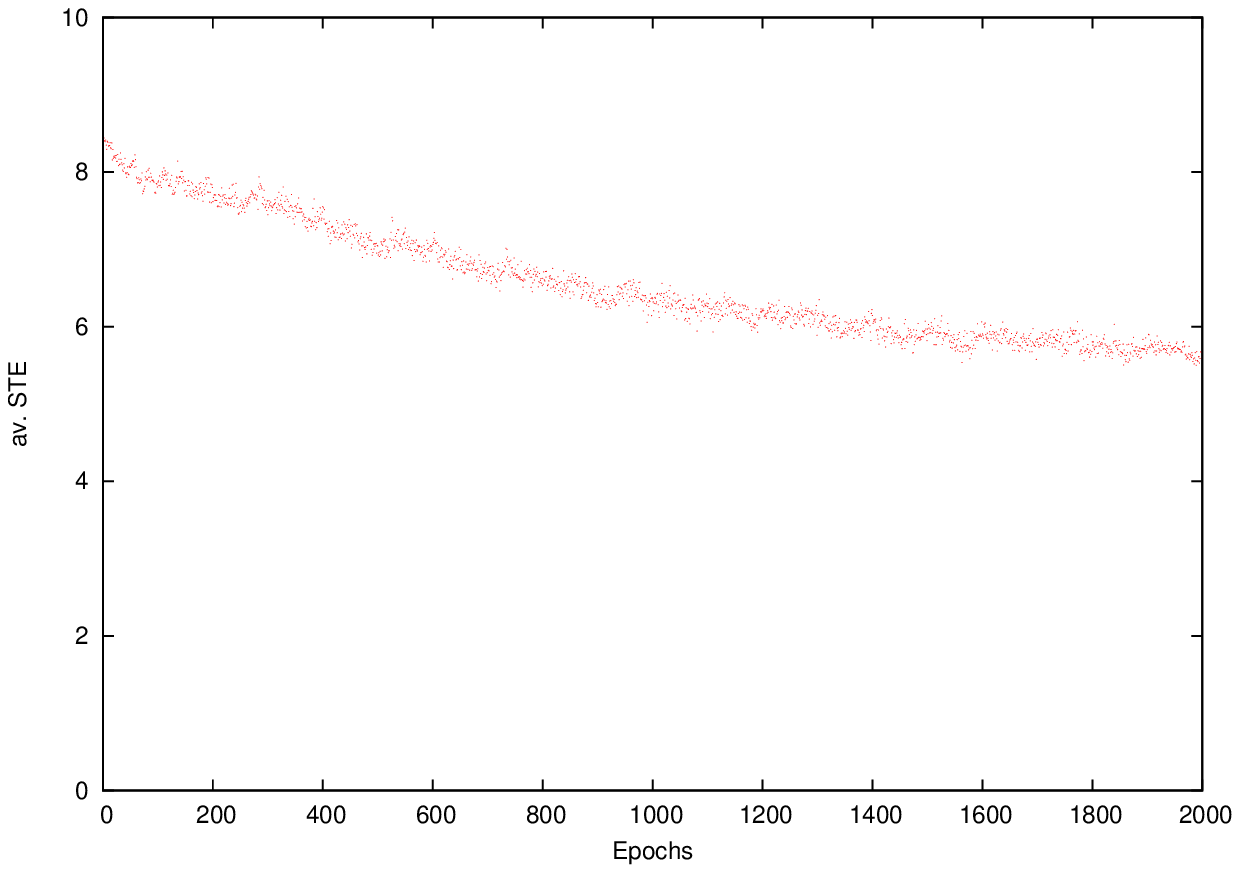}}
  \subfloat[\ac{STE} for \AND]{\epsfig{width=0.45\hsize,file=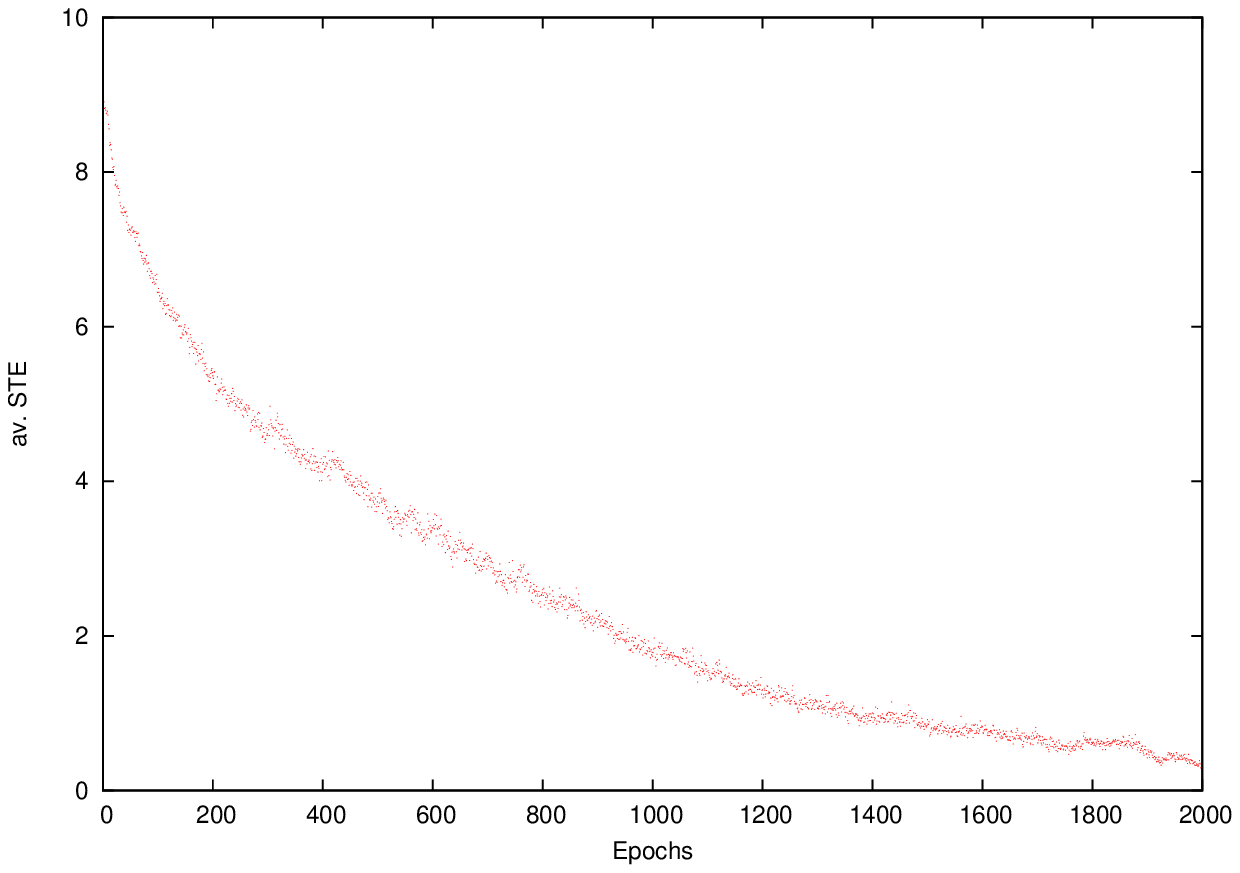}}
  \\
%
  \subfloat[\ac{LE} for \XOR]{\epsfig{width=0.45\hsize,file=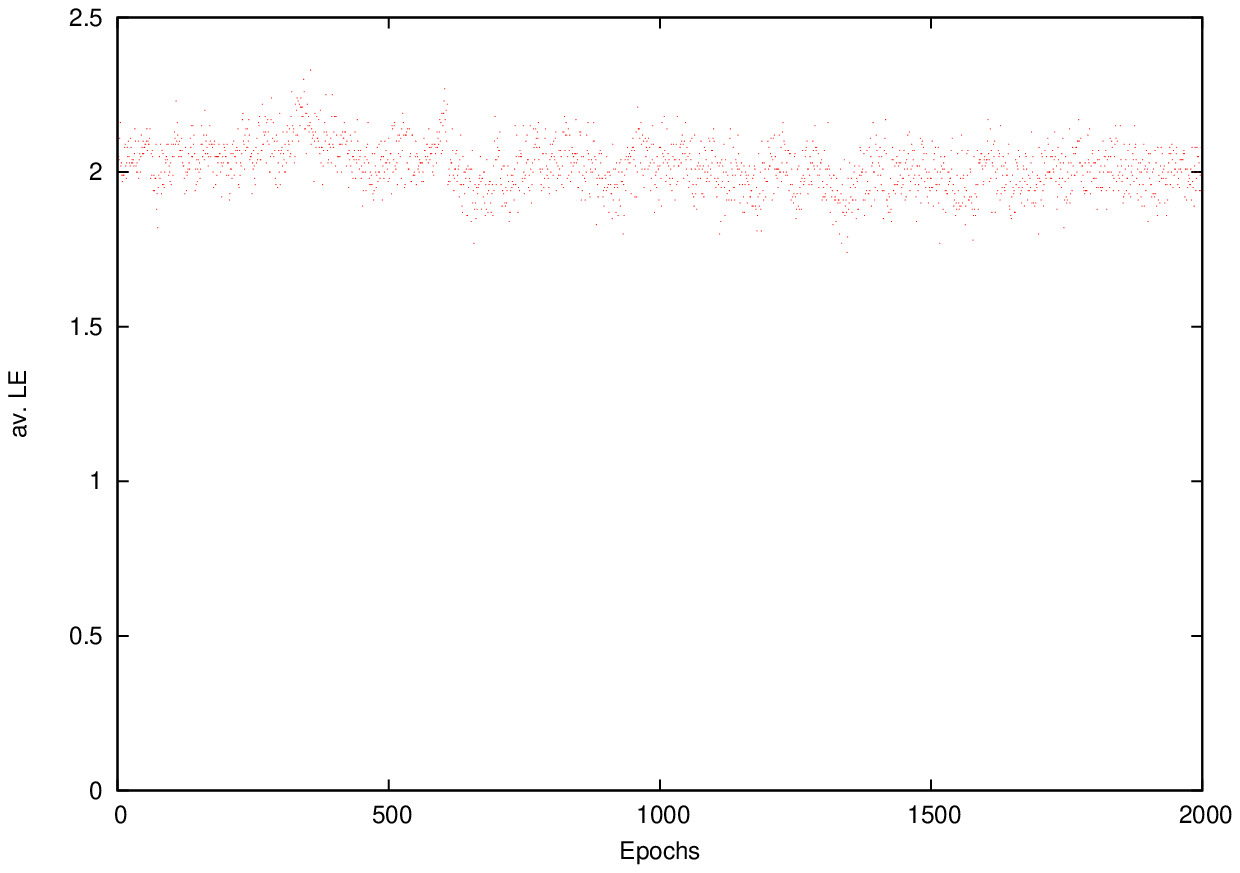}}
  \subfloat[\ac{LE} for \AND]{\epsfig{width=0.45\hsize,file=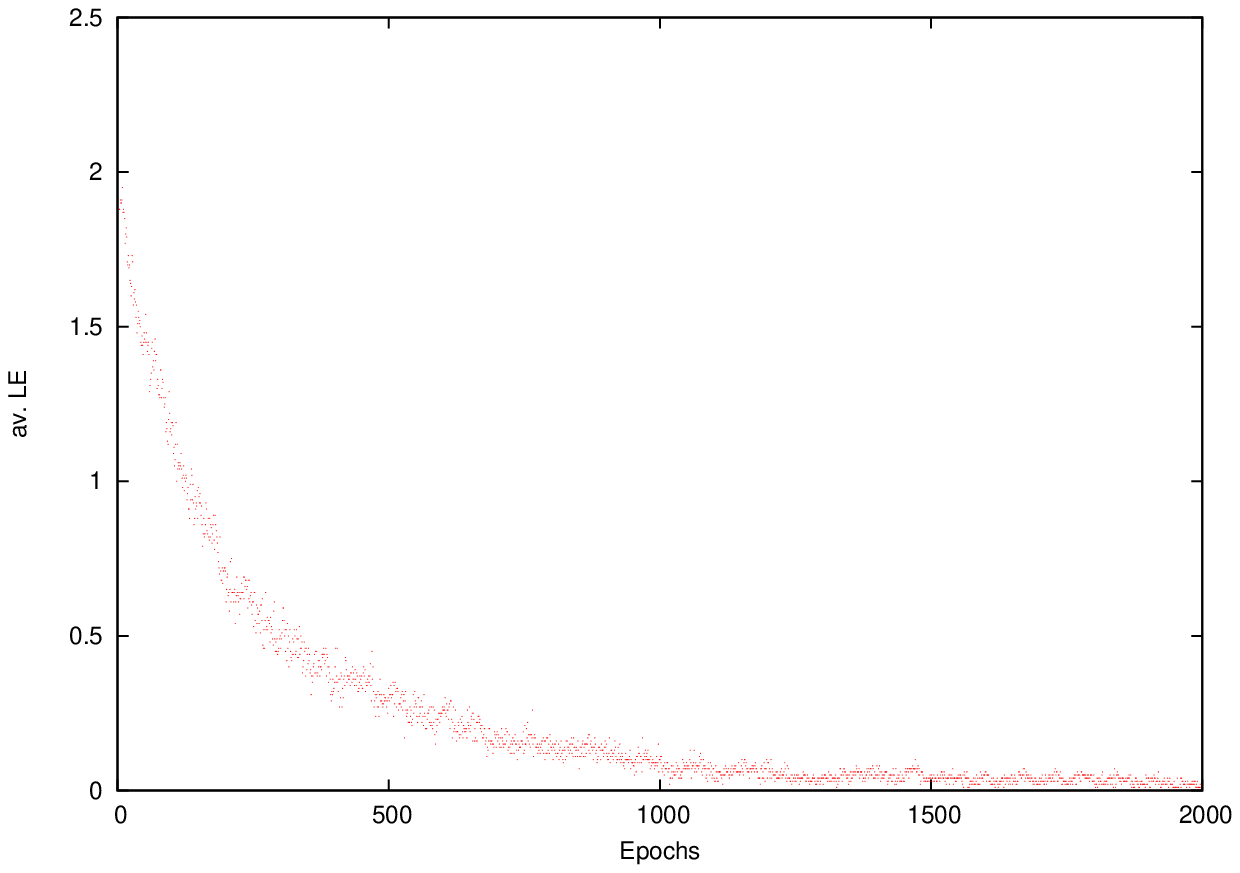}}
  \\
  \caption{Average \ac{STE} and \ac{LE} for \XOR{} and \AND{} for
    the two-layer network with 20 input neurons and without hidden
    layer. The \ac{STE} graphs are clipped at 10 as this
    error is only exceeded for the first 20 epochs as discussed in the
    main text.} \label{fig:av_20_ste} \label{fig:av_20_le}
\end{figure}

\begin{table}
  \small
  \centering
  \caption{Mean \ac{STE} and \ac{LE} averaged over epochs 900--999 and
  1900-1999 and 100 networks for each condition. Most significant digit of error of the mean shown in ().}   
  \label{tab:results}

  \subfloat[12 inputs, 20 hidden neurons]{%
    \begin{tabular}{|l|l|c|c|}
      \hline
      Operation & Error & 900-999 & 1900-1999 \\
      \hline
      \multirow{2}{*}{\AND{}} & \ac{STE} &	 3.37(2)  &	 2.35(3) \\
                              & \ac{LE}  &	0.170(3)  &  	0.076(2) \\ \hline
      \multirow{2}{*}{\PROJ}  & \ac{STE} & 	3.84(2)   &  	2.83(3) \\ 
                              & \ac{LE}  & 	0.230(4) &	0.149(3) \\ \hline
      \multirow{2}{*}{\TRUE}  & \ac{STE} & 	3.32(2) & 	2.55(3) \\ 
                              & \ac{LE}  & 	0.161(3) &	0.078(3) \\ \hline
      \multirow{2}{*}{\XOR}   & \ac{STE} &  	3.55(2) 	&3.08(3) \\ 
                              & \ac{LE}  &	0.200(4)  &	0.157(2) \\ \hline
    \end{tabular}
  }   
  \subfloat[20 inputs, no hidden layer]{%
    \begin{tabular}{|l|l|c|c|}
      \hline
      Operation & Error & 900-999 & 1900-1999 \\
      \hline
      \multirow{2}{*}{\AND} & \ac{STE} & 1.98(1) 	&0.41(1) \\ 
                            & \ac{LE}  & 0.104(2) 	&0.022(1) \\  \hline
      \multirow{2}{*}{\PROJ}  & \ac{STE} &1.29(1) 	&0.37(1) \\ 
                              & \ac{LE} &0.047(2) 	&0.007(1) \\ \hline
      \multirow{2}{*}{\TRUE}  & \ac{STE} &0.570(9) 	&0.084(2) \\ 
                              & \ac{LE}	& 0.010(1)	&0 \\ \hline
      \multirow{2}{*}{\XOR}   & \ac{STE} & 6.39(1) 	&5.70(1) \\ 
                              & \ac{LE} &2.012(8) 	&1.994(7) \\ \hline
    \end{tabular}
  }   
  \\
  \subfloat[Control case, 12 inputs, 12 hidden neurons]{%
    \begin{tabular}{|l|l|c|c|}
      \hline
      Operation & Error & 900-999 & 1900-1999 \\
      \hline
      \multirow{2}{*}{\AND} & \ac{STE} & 6.26(2) &5.52(2) \\
                            & \ac{LE}  & 0.75(5) &0.469(4) \\ \hline
      \multirow{2}{*}{\PROJ}  & \ac{STE} &6.03(2) &5.60(3) \\ 
                              & \ac{LE}	&0.612(5) &0.390(5) \\ \hline
      \multirow{2}{*}{\TRUE}  & \ac{STE} &6.34(2) &5.29(3) \\ 
                              & \ac{LE} &0.641(6) &0.371(6) \\ \hline
      \multirow{2}{*}{\XOR}   & \ac{STE} & 6.56(2) &5.978(3) \\
                              & \ac{LE} & 0.761(5) & 0.518(5) \\ \hline
    \end{tabular}
  }
  \subfloat[Control Case: 12 inputs, no hidden layer]{%
    \begin{tabular}{|l|l|c|c|}
      \hline
      Operation & Error & 900-999 & 1900-1999 \\
      \hline
      \multirow{2}{*}{\AND} & \ac{STE}  & 6.69(1) &5.90(1) \\ 
                            & \ac{LE}   & 0.843(5) & 0.648(5) \\ \hline
      \multirow{2}{*}{\PROJ}  & \ac{STE}	&4.70(1) 	&3.83(1) \\ 
                              &  \ac{LE}	&0.377(4) 	&0.261(3) \\ \hline
      \multirow{2}{*}{\TRUE}  & \ac{STE} 	&5.35(1) 	&3.78(1) \\ 
                              &  \ac{LE} 	&0.453(4) 	&0.179(3) \\     \hline
      \multirow{2}{*}{\XOR}   & \ac{STE} 	&9.16(2) 	&8.88(1) \\ 
                              & \ac{LE} 	&1.978(8) 	&1.968(7) \\ \hline
    \end{tabular}
  }      
\end{table}

\subsection{Discussion}  

Figures~\ref{fig:av_12_20_ste}--\ref{fig:av_20_le} present average
learning curves for \ac{STE} and \ac{LE} for the two network
configurations for logical operations \XOR{} and \AND{} averaged over
100 runs. Learning curves for \J{} and \TRUE{} are very similar to
\AND{} for each network and are therefore not shown.
\par
The course for \ac{STE} curves is similar for all cases: From about
epoch~20 the \ac{STE} errors are below 10 and decay first relatively
steeply and then slower. However the \ac{STE} for the three-layer
networks starts with a low error value in the 
first epochs which rises steeply to very high values up to 400 and then
rapidly decays to values below 10 in the first 20
epochs (not shown, \ac{STE} graphs clipped at 10). The reason for the marked peak in three-layer networks only is
that initial weights were chosen so that no output spikes are generated at all. As weights increase by
scaling \prettyref{eq:adj} and as more output spikes are generated, the
errors increase until most of them are removed again via
\prettyref{eq:dwa}. For \ac{LE} the picture is similar, however
testing the interpretation of 
the output, \ac{LE} starts always at the random level 2, perhaps
slightly increasing beyond that around epoch~20 and then decreases
 to lower values (with the exception of \XOR{} for the two-layer
 networks, see below). All in all, errors decrease in a way similar to many other supervised network learning
algorithms. \par
The \XOR{} problem in the three-layer network reaches level after
about 1500 epochs and does not decay after 
that, see \prettyref{fig:av_12_20_ste}(a), while the \ac{STE} learning curve for
\XOR{} in the two-layer networks is still decreasing after 2000
epochs, see \prettyref{fig:av_20_ste}(a). Additional simulations for up to 10000 epochs however confirmed
that the \ac{STE} reaches a minimum after about 3000 epochs for the
two-layer networks, and that \ac{LE} does not change at all, but
fluctuates around the random performance value $2$ throughout all epochs.
\par
It may be seen that all curves are very rough in nature and this is
discussed below. Tables \ref{tab:results}(a) and (b) summarise the simulation
results in terms of \ac{STE} and \ac{LE} averaged over
all 100 networks for a given configuration and the 100 epochs from 900--999 and
1900--1999 respectively. 
\par
The amount of activation that the output neuron can
get from its predecessor layer is similar for the three (with 20 neurons in the
hidden layer) and the two-layer networks (with 20 neurons in the input layer)
and a comparison between them is interesting. We discuss the different
logical operations at this point.  The two-layer network is best
except for the \XOR{} operation, and its \ac{LE} performance on \XOR{}
stays throughout training at a random level. In other words,
the networks without hidden layer are not able to learn the \XOR{}
problem reliably. 
 \par
The \XOR{} operation reaches an \ac{LE} value of between 0.157
and 0.200 for the three-layer networks. If we assume that networks
with $\ac{LE} \neq 0$ have at most $\ac{LE} = 1$, then this indicates
that more than a fraction of $0.843 = 1 - 0.157$ of these networks
produces no logical error on average for any epoch from 1900-1999. In
other words, the three-layer networks can learn the \XOR{} problem,
although not with a reliability fit for technical purposes. 
\par
As to the other logical operations, \TRUE{} and \PROJ{} do best in two
layer networks. \AND{} and \XOR{} are the operations where information
from the two inputs $J_0$ and $J_1$ needs to be combined, and they are learned with a
somewhat higher \ac{STE} and \ac{LE} error rate.  For the three-layer network,
\PROJ{} is hardest to learn. With a large hidden layer that mixes
spikes from both input banks it might be more difficult to find enough
spikes that convey information about only one of the inputs banks.
\par
Overall relative differences between logical operations are lower for
the three-layer networks. Hence the difficulties to match a spike train 
outweigh the difficulty to perform the computation.

\subsection{Error Roughness}

\begin{figure}
  \centering
  \subfloat[\ac{STE} for all epochs, clipped at 20.]{%
    \epsfig{width=0.45\hsize,file=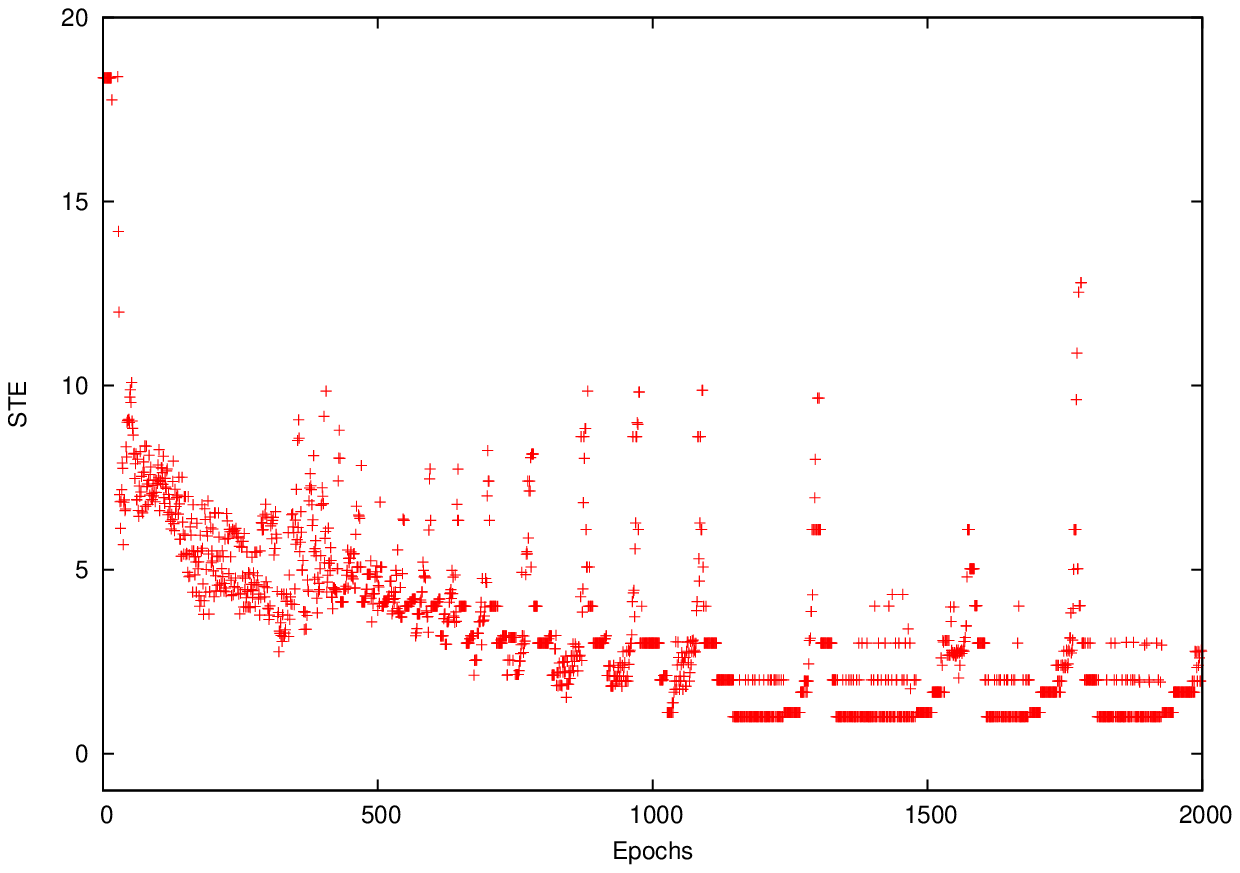}
    }
    \subfloat[\ac{STE}. Enlargement of (a) for epochs 750--1500.]{%
      \epsfig{width=0.45\hsize,file=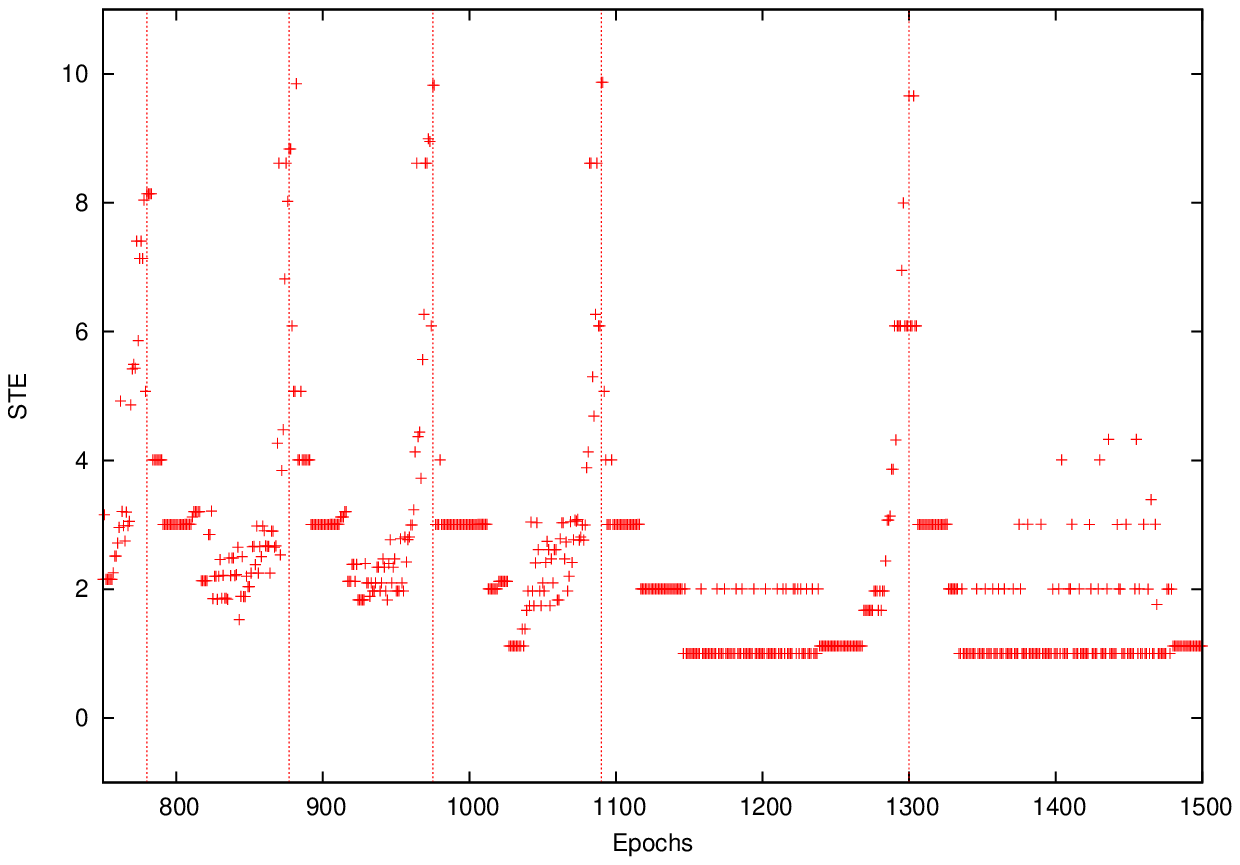}}  
    \\
    \subfloat[\ac{LE} for all epochs.]{%
      \epsfig{width=0.45\hsize,file=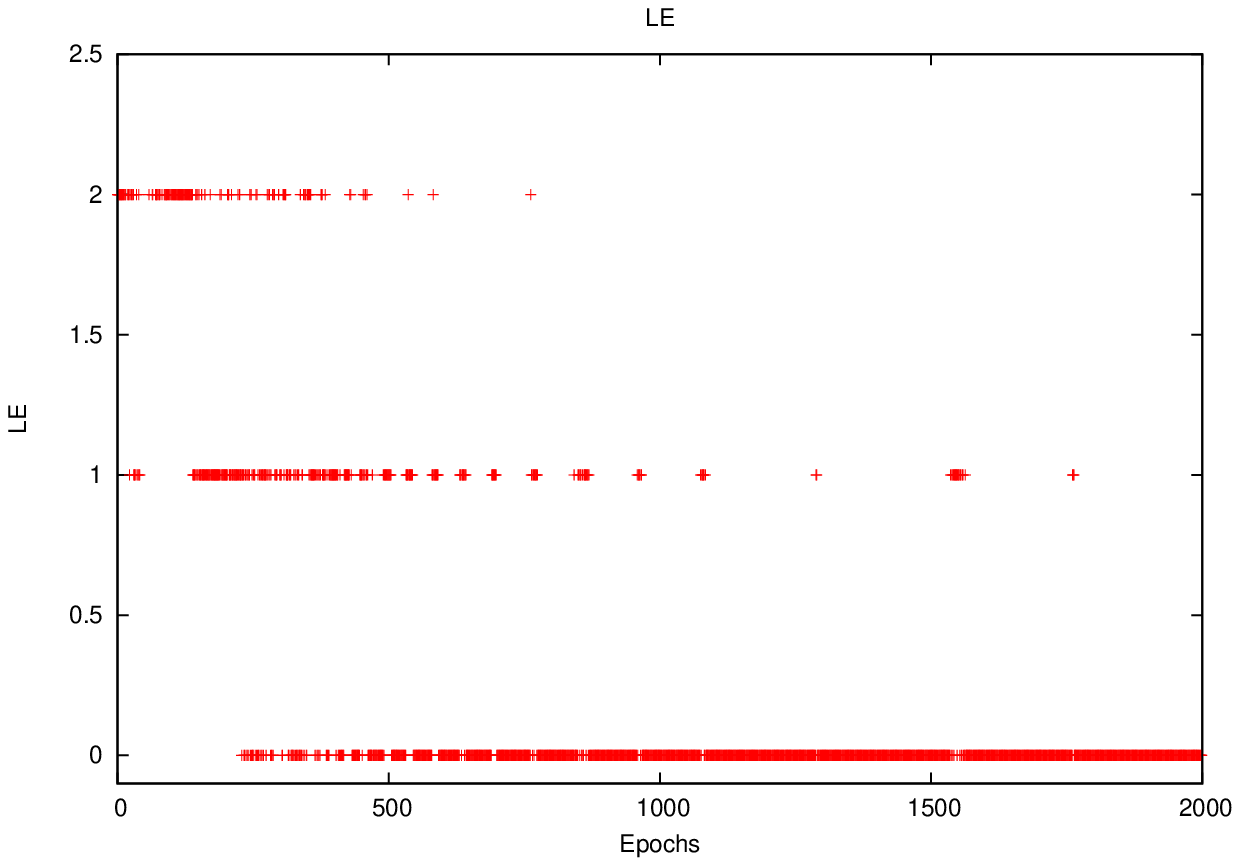}
    }
    \subfloat[\ac{LE}. Enlargement of (c) for epochs
    750--1500.]{\epsfig{width=0.45\hsize,file=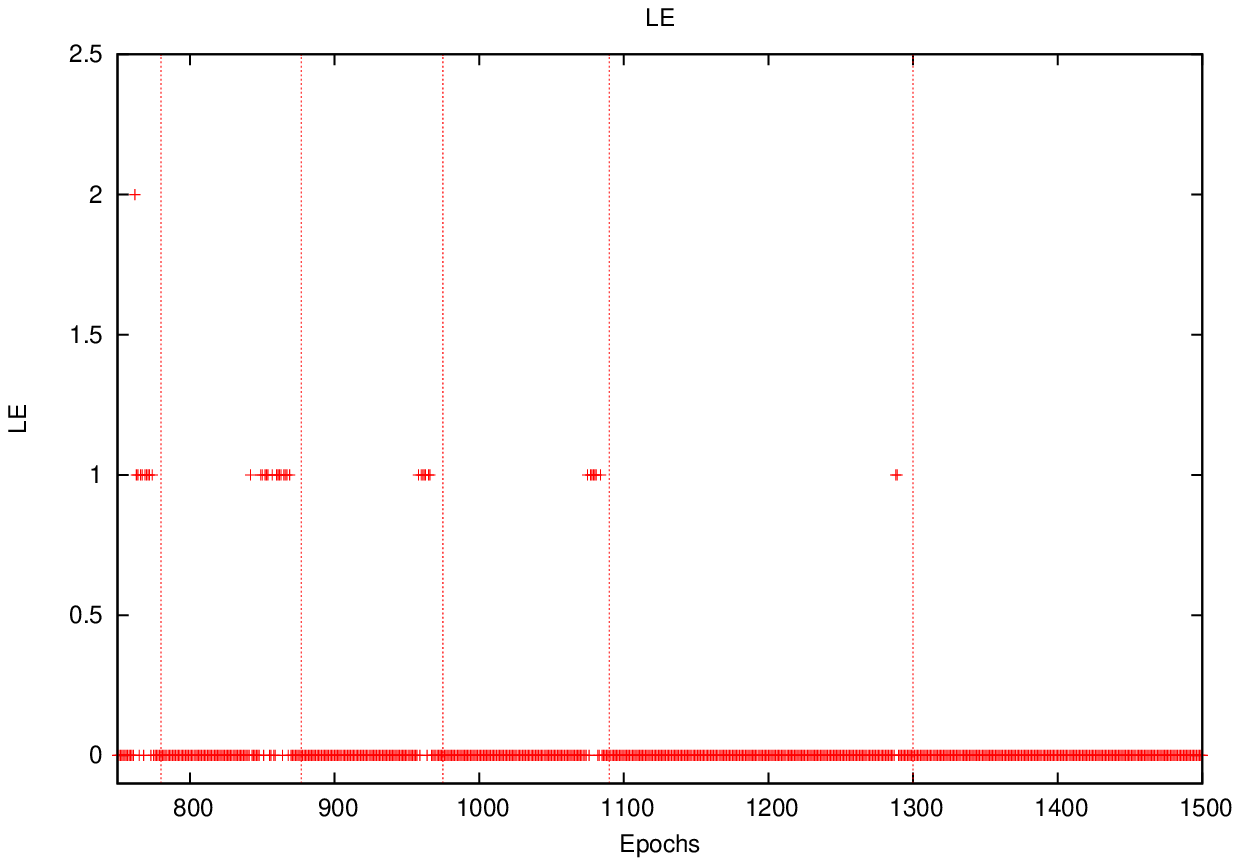}}   
    \\
    \subfloat[Spike Trains for epochs 750--1500.]{%
      \epsfig{width=0.45\hsize,file=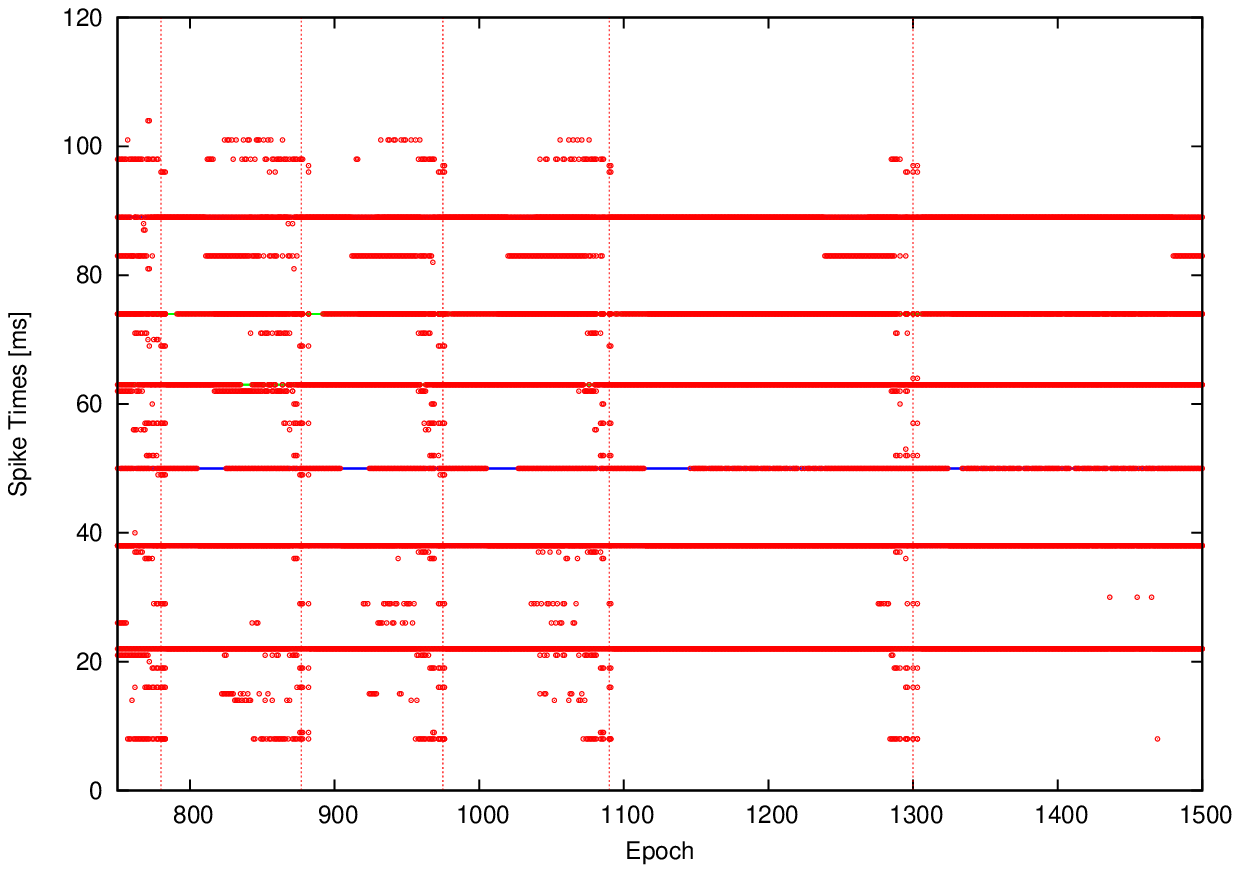}} 
    \caption{Typical individual learning curves and spike train evolution for a
      single network with 12 inputs and 20 hidden neurons trained on
      \XOR. (a)/(c) The network is generally performing well from about 750
      epochs, however it loses a solution that is close to the target
      spike trains for short times, indicated by the
      peaks of \ac{STE}. However \ac{LE} is less effected.  
      (b)/(d) Enlargement of (a)/(c) for epochs 750--1500. Vertical
      dotted lines indicate the peaks of \ac{STE}. \ac{LE} deviates
      from 0 immediately before the \ac{STE} peaks and then returns to 0.
      Towards the end of the cascade of peaks the networks settles
      into a closer approximation of the target spike trains with
      \ac{STE} frequently at about $1.5$. 
      (e) Spike times evolution for epochs 750--1500: The actual output spike times are
      represented as rings. Desired spike times are represented as the
      solid horizontal lines. These are covered heavily with actual spikes that hit the right
      spike time. This graph overlays actual and desired output spike
      times for \emph{all} \XOR{} input-output patterns 000,011,101,110. That all desired spikes time seem to be hit,
      does not imply that any single output train hits all target
      times.  Vertical dotted lines again indicate the epochs with
      \ac{STE} peaks. These are correlated with a reorganisation of
      spike time patterns. 
}    \label{fig:individual}
\end{figure}

Despite averaging over 100 networks, it is obvious that the
learning curves are not as smooth as for rate neuron learning
algorithms. This roughness stems from that of individual learning
curves, see  \prettyref{fig:individual} for a typical successful
network trained on the \textsc{XOR} task. Individual networks occasionally lose a 
good solution and refind it later. This can lead to large changes in 
\ac{STE}. However
changes in \ac{LE} might not be as pronounced as in \ac{STE}, see
\prettyref{fig:individual}(c). 
These abrupt changes are an effect of the discontinuous nature of
spike events, where relatively large discrete jumps in the errors for
individual networks are expected when weights are slightly changed but
an additional output spike is created or disappears. 

\subsection{Further Control Cases}

We also ran two further network configurations as control cases,
namely a three-layer network still with 12 inputs, but only 12 neurons
in the hidden layer, and a two-layer network with only 12
inputs. Learning curves were qualitatively similar to the other
networks, but performance (see \prettyref{tab:results}(c) and (d)) was worse than
their counterparts with bigger hidden or input layer. 
\par
That the three-layer network with 20 hidden neurons and 12
inputs performs better on all logical operation than the two-layer
network with 12 inputs, demonstrates that, as for rate neuron
networks, a hidden layer is useful to preprocess and mix inputs even
though the total information fed into the networks is the same.

\section{Conclusion} \label{sec:conclusion}

The present simulations -- to our knowledge for the first time --
present an example of supervised learning in layered spiking neural
networks where \emph{inputs and outputs are encoded as spike
  trains} of multiple spikes. It extends and builds on other
supervised learning algorithms for spiking neural networks like
\rsm{} and SpikeProp. Restrictions on spike patterns in SpikeProp
and its extensions (one latency-coded output spike) are more severe than in
the present simulations (three timed output spikes).
\rsm{} has only been used on either single neurons or on read-outs
of \ac{LSM}, nor has it been attempted to implement a simple but
non-trivial computation like the \XOR{} operation. SpikeProp suffers from silent neurons in the hidden
layer for which no error signals can be obtained. We sidestep a similar
problem by scaling weights multiplicatively so that firing rates are
kept within a specified range. 
\par
Our results indicate that on average more than 80\% of the three-layer
networks in any one of the final 100 epochs compute the not linearly
separable \XOR{} operation correctly while two-layer networks do
not. This extends a similar observation for layered networks of \emph{rate}
neurons \citep{minsky:88}. However the roughness of the learning
curves suggests that networks frequently lose and refind a good
solution.
\par
\rsm{} as applied to layered networks is certainly not reliable
enough for technical purposes or even for information 
processing in the nervous system. However we have so far only
considered a single output spike train. If a single neuron and its
spike train are individually not reliable,  they may be as an
ensemble. It has been observed that neurons driven with the same inputs can
be trained to produce different spike trains \citep{kasinski:05}. It
is therefore possible that a bank of output neurons driven
from the same hidden layer, but producing different spike trains for
the same logical value, can be trained successfully, as incoming weights
of the output neurons and their targets are independent. This ensemble would be more 
reliable to represent the \true{} or \false{} output than any neuron on its
own \citep{polikar:06}. Multiple output spike trains  mirror using banks of inputs,
too. In addition, this is also more realistic as in nature neurons act collectively to encode
information and omission or addition of single spikes does not seem to
be critical \citep{gerstein:01, izhikevich:07}.  
\par
While it is often clear that a given network structure is
Turing-equivalent, it is less clear what computations can be learnt and
how they are implemented on a given network in a natural way. 
There has been a successful stream of research to analyse what
computational representations a rate network evolves for a given
computational problem \citep{boden:00b, rodriguez:01,
  gruening:05}. In this spirit, we believe it is now time to explore spiking
neural networks and their computational capabilities. 
\par



\subsection*{Acknowledgements}

  The authors thank Yaochu Jin, Scott Notley and Susanne Schindler
  for comments on an earlier version of this article, and Susanne also
  for support with \prettyref{fig:network}. AG also personally thanks
  Susanne Schindler (and the AP) for her invaluable company without
  which this work would not have been. 
  \par
  AG's part of the work was conducted under the Engineering and Physical
  Sciences Research Council (UK) grant EP/I014934/1.

\bibliographystyle{spbasic}

\bibliography{doktor,publications}

\end{document}